\documentclass[runningheads]{llncs}
\usepackage[T1]{fontenc}
\usepackage{graphicx}
\usepackage{wrapfig}
\graphicspath{ {./pic/} }
\usepackage{amsmath}
\usepackage{amsfonts}
\usepackage{hyperref}

\usepackage[titlenumbered,ruled]{algorithm2e}
\SetKwInOut{Input}{Input}
\SetKwInOut{Output}{Output}

\usepackage{ifthen}
\usepackage{xcolor}

\begin{document}
\title{Reinforcement Learning with Success Induced Task Prioritization}
%
%

\author{Maria Nesterova\inst{1} \and Alexey Skrynnik\inst{1, 2, 3} \and Aleksandr Panov\inst{2,3}}
\authorrunning{M. Nesterova et al.}

\institute{Moscow Institute of Physics and Technology, Moscow, Russia\\
\and AIRI, Moscow, Russia \\ 
\and Federal Research Center ``Computer Science and Control'' of the Russian Academy of Sciences, Moscow, Russia \\
\email{nesterova.mi@phystech.edu}}

\maketitle  
\begin{abstract}
Many challenging reinforcement learning (RL) problems require designing a distribution of tasks that can be applied to train effective policies. This distribution of tasks can be specified by the curriculum. A curriculum is meant to improve the results of learning and accelerate it. We introduce Success Induced Task Prioritization (SITP), a framework for automatic curriculum learning, where a task sequence is created based on the success rate of each task. In this setting, each task is an algorithmically created environment instance with a unique configuration. The algorithm selects the order of tasks that provide the fastest learning for agents. The probability of selecting any of the tasks for the next stage of learning is determined by evaluating its performance score in previous stages. Experiments were carried out in the Partially Observable Grid Environment for Multiple Agents (POGEMA) and Procgen benchmark. We demonstrate that SITP matches or surpasses the results of other curriculum design methods. Our method can be implemented with handful of minor modifications to any standard RL framework and provides useful prioritization with minimal computational overhead.

\end{abstract}

\section{Introduction}

\begin{figure}[ht]
    \begin{center}
        \includegraphics[width=1\linewidth]{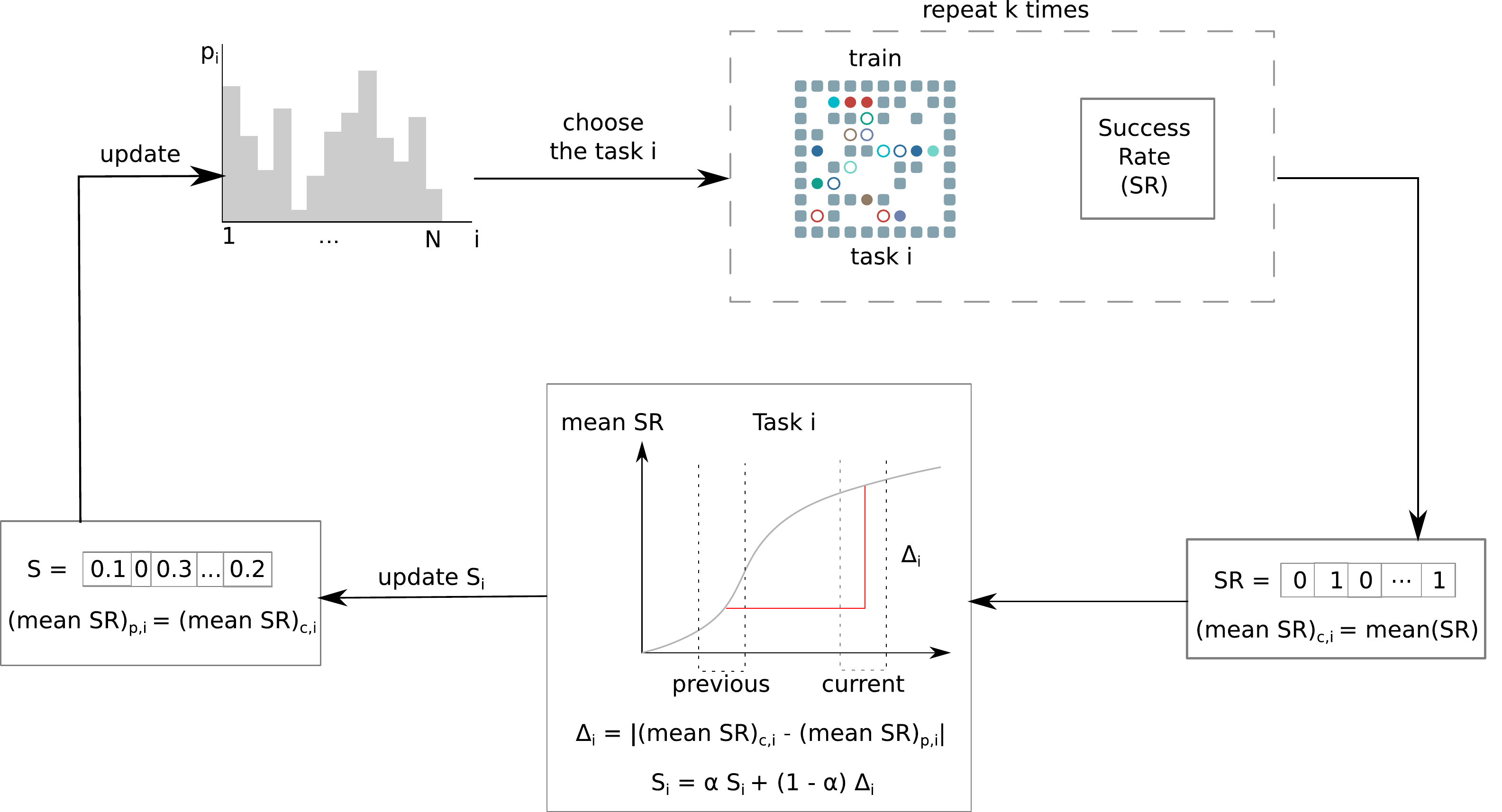}
    \end{center} 
    \caption{Overview of SITP method. The probability of selecting task $i$ depends on how much earlier the mean Success Rate (SR) increased after training on task $i$, $k$ times.}
    \label{pic:meta-sitp}
\end{figure}

In numerous complex Reinforcement Learning (RL) problems, the agent must master a number of tasks. That number of tasks may be explicitly defined by the environment authors or may be implicit. E.g. in case then the environment is procedurally generated, and the agent is given only a few world instances in which it can learn.
The proper order can, on the one hand, speed up the agent's learning process, and on the other hand, prevent him from catastrophic forgetting.
Training using task sequencing order is called curriculum learning~\cite{Narvekar2020CurriculumLF}.
More formally, the curriculum learning is a method of optimization of the order in which experience is accumulated by the agent, so as to improve performance or training speed on a set of final tasks.  

There are many ways of creating a curriculum. One of the possible ways of constructing a curriculum is to arrange the tasks by difficulty. The second option is to choose a sequence of tasks according to a certain distribution, using additional information received during training~\cite{level-replay,teacher-student}. The third option is to create a teacher, whose goal is to gradually increase the complexity of the tasks, while the teacher itself learns to create them ~\cite{chen2021variational}. For example, the teacher can generate obstacles on the map~\cite{dennis2020emergent}. It is also possible to create an implicit curriculum based on competition or interaction between agents~\cite{level-replay,teacher-student}. 

In this paper, we introduce Success Induced Task Prioritization (SITP)\footnote[1]{Our code is available at \href{https://github.com/nortem/sitp}{https://github.com/nortem/sitp}}, illustrated in Fig.~\ref{pic:meta-sitp}, a new method for task sequencing. The main concept for this method is a binary metric, Success Rate (SR), which shows whether the task was completed on the episode or not. During training, the method updates scores estimating each task's learning potential, basing on mean SR for this task. Then this method selects the next training task from a distribution derived from a normalization procedure over these task scores. Our method also does not apply any external, predefined ordering of tasks by difficulty or other criteria, but instead derives task scores dynamically during training based on SR. It is assumed that tasks which get the highest rate of selection are the ones on which the agents learn the fastest, or that those tasks are the hardest to complete.

We consider the SITP method in the application for the Multi-agent Pathfinding (MAPF) problem, which is based on POGEMA environment~\cite{pogema}. The MAPF problem is that several agents must go from their starting positions to the goals without colliding with obstacles or other agents. In this problem domain, SR is defined as an indicator of whether all agents have reached their goals or not. We argue that such simple information is enough to improve learning results. Moreover, we provide additional experiments with a well-known Procgen benchmark~\cite{cobbe2020leveraging}, showing the applicability of the SITP approach to a wide class of RL problems.

The rest of the paper is organized as follows. Section 2 provides a brief overview of related works. Section 3 describes the background behind RL. Section 4 introduces SITP method. Section 5 is devoted for experimental study of presented method and comparison with other approaches. In the conclusion we discuss obtained results.

\section{Related Work}

The central problem of creating a curriculum is the definition of a sequence of tasks or the generation of tasks automatically (without human intervention).  Several papers address the last one, showing that it is possible to generate a curriculum automatically by confronting several agents~\cite{baker-autocurricula,Narvekar2020CurriculumLF}. A similar idea assumes the sequential interaction of two agents~\cite{sukhbaatar-intrinsic}, where one sets the task, and the other solves it. 

Separately, one can consider the methods that create the curriculum for a set of already known tasks.
The Prioritized Level Replay (PLR)~\cite{level-replay} provides a curriculum scheme automatically using additional information collected during training. The main idea is to define a priority for each task using some scoring scheme. The authors propose to accumulate L1 General Advantage Estimation (GAE) during training for each task and sample new ones based on that scores. This score shows whether it is promising to train on this task in the future. On the other hand, the Teacher-Student Curriculum Learning (TSCL)~\cite{teacher-student}  provides several ways to estimate the prospects of a task using the learning progress curve. Despite the method's conceptual simplicity, there may be difficulties with the formation of a set of tasks or the choice of a metric for that tasks.

The tasks could be created in process of training an additional agent, as in the self-play algorithm \cite{sukhbaatar-intrinsic}. This algorithm is focused on two kinds of environments: reversible environments and environments that can be reset. An automatic learning program is created based on the interaction of two agents: Alice and Bob. Alice will “propose” the task by doing a sequence of actions and then Bob must undo or repeat them, respectively. Authors argue that this way of creating a training program is effective in different types of environments.

\section{Background}

The reinforcement learning problem is to train an agent (or several agents) how to act in a certain environment so as to maximize some scalar reward signal. The process of interaction between the agent and the environment is modeled by the partially observable Markov decision process (POMDP), which is a variant of the (plain) Markov decision process (MDP).
Partially Observable Markov Decision Process (POMDP) is a tuple  \(\langle \mathcal{S} , A, O, \mathcal{T} , p, r, \mathcal{I}, p_o, \gamma \rangle\), where: \(\mathcal{S}\) -- set of states of the environment, \(A\) -- set of available actions, \(O\) -- set of observations, \(\mathcal{T} : \mathcal{S} \times A \rightarrow \mathcal{S}\) -- transition function, \(p(s' | s, a)\) -- probability of transition to state \(s'\) from state \(s\) under the action \(a\), \(r: \mathcal{S} \times A \rightarrow \mathbb{R}\) -- reward function, \(\mathcal{I}: \mathcal{S} \rightarrow O\) -- observation function, \(p_o(o | s', a)\) -- probability to get observation \(o\) if the state transitioned to \(s'\) under the action \(a\), \(\gamma \in [0, 1]\) -- discount factor. At each timestep $t$, the agent chooses its action $a_t$ based on the policy $\pi (a|s): A \times \mathcal{S} \rightarrow [0, 1]$ and receives the reward $r_t$. The goal of the agent is to learn an optimal policy $\pi^*$, which maximizes the expected return.

In this paper we apply the Proximal Policy Optimization (PPO)~\cite{ar:schulman_PPO} algorithm to find that policy. PPO has proven itself as a highly robust approach for many tasks, including multi-agent scenarios~\cite{ar:IPPO}, large-scale training~\cite{berner2019dota}, and even fine-tuning other policies ~\cite{baker-minecraft}. PPO is a variation of advantage actor-critic, which leans on clipping in the objective function to penalize the new policy for getting far from the old one.

\section{Success Induced Task Prioritization}

In this section, we present Success Induced Task Prioritization (SITP), an algorithm for selecting the next task for learning by prioritizing tasks basing on previous learning results.
SITP assumes that a finite set of $N$ tasks $T = [T_1, \dots, T_N]$ is defined in the environment. A task is an algorithmically created environment instance with unique configuration. Task configuration is a broad term, several instances of the environment, united by one common property, can be located inside the same task. 
We assume that if the agent learns tasks in a certain order, then this will contribute to improve results of learning and accelerate it. 

A curriculum $C$ is a directed acyclic graph that establishes a partial order in which tasks should be trained~\cite{Narvekar2020CurriculumLF}. Linear sequence is the simplest and most common structure for a curriculum. A curriculum can be created online, where the task order is determined dynamically based on the agent's learning progress.

An important feature of our algorithm is the application of the success rate (SR). SR is a binary score that is utilized to measure the success of training on a single episode. There are several ways to set the metric. For example, if the environment has a specific goal that the agent must achieve, then SR = 1 if this goal is achieved, 0 otherwise. If there is no specific goal, then SR can be determined using the reward for the episode (in case the reward is bigger than a certain threshold, then SR = 1, otherwise 0). The concept of a mean SR is introduced to evaluate the effectiveness of an agent. Mean SR shows the percentage of successfully completed episodes.

We assume that with proper task sequencing the speed of learning and success rate could be increased. The target task is the task on which mean SR is meant to improve. In cases where  more than one task is being considered to be the target task, the curriculum is supposed to improve the mean SR of all of them. All of the methods described below are meant to be used for multiple target tasks.

Assume, $N$ tasks $T = [T_1, \dots, T_N]$ and number of iterations $M$ of the algorithm are given. Let $T_i$ be a task, $S_i$ a score, obtained as a result of training on this task, $p_i$ a probability of the task being selected on the next training iteration:

\begin{equation}
    p_i = \dfrac{\exp(S_i)}{\sum\limits_{j = 1}^N \exp(S_j)}, \; i = \overline{1, N}.
\end{equation}

In such a way, probability distribution $p = [p_1, \dots, p_N]$ is constructed. It gives priority to tasks depending on scores $S = [S_1, \dots, S_N]$. Initially the tasks are sequenced with equal probability. General idea of our method illustrated in Fig.~\ref{pic:meta-sitp} and specified in Algorithm~\ref{meta_alg}.

\begin{algorithm}[H] \label{meta_alg}
    \Input{Tasks $T$, $N$ -- number of tasks}
    Initialize agents learning algorithm\\
    Initialize probability distribution $p \gets [\frac{1}{N}, \dots, \frac{1}{N}]$\\
    Initialize scores $S$ 

    \For{$t = 1, \dots, M$}{
        Choose task $T_i$ based on $p$\\
        Train agents using task $T_i$ based on the evaluation method and observe $score$\\
        Update score $S[i] \gets score$\\
        Update probability distribution $p$
    }
    
    \caption{Training loop with SITP}
\end{algorithm}

Depending on the way of evaluating S, different variations of an algorithm are possible. SITP is based on the idea that if one of the tasks increases mean SR more than others, then this task is better to be leaning on. We presume that on such task SR will keep increasing. In the same manner, forgetting of the task is taken into account. If mean SR is decreasing on a certain task, then this task should be run again. In such a way, evaluation of $S_i$ depends on the absolute value of mean SR's change. The bigger the effect on mean SR from the $T_i$ task, the bigger the chance of it being selected. Thus, at first the task $T_i$ is selected, basing on probability distribution $p$. Then the agents learn and $S_i$ scores for $T_i$ tasks are evaluated. Agents train for  $k$ episodes on the selected task  $T_i$, get a SR for each episode, and then mean SR is calculated.  $SR^i_{old}$- mean SR of previous learning stage on  $T_i$ task. Initially $SR^i_{old}$ equals 0.  The mean SR changes during the training of agents by $\Delta_i = |SR^i_{new} - SR^i_{old}|$. A moving average also should be used for smoothing out short-term fluctuations.  The final score is calculated: $$S_i = \alpha S_i  + (1-\alpha) | SR_{new}^i - SR_{old}^i |,$$ where $\alpha$ is smoothing coefficient. Additionally, a condition is established that if $SR^i_{new}$ exceeds a certain threshold $max\_SR$, then it is considered that agents are good enough at solving this problem and it should be chosen less frequently. This is controlled by the $min\_score$ number. The procedure for constructing the estimate $S_i$ is shown in Algorithm \ref{alg_2}.

\begin{algorithm}[H] \label{alg_2}
    \Input{$T_i$ -- selected task, $k$ -- number of episodes, $SR^i_{old}$ -- mean SR for the previous stage, $max\_SR$ -- the maximum mean SR, when this mean SR is reached, then the task should be selected less frequently, $min\_score$ -- the number by which the task is selected less frequently, $\alpha$ -- smoothing coefficient.}
    \Output{Score $S_i$ for $T_i$, $SR^i_{old}$}

    Initialize $SR$ for $k$ episodes $C \gets [0, \dots, 0]$\\
    \If{first stage for $T_i$}{
        Initialize $SR^i_{old} \gets 0$ 
    }

    \For{$\tau = 1, \dots, k$}{
            Train agents using task $T_i$ and observe $SR$ score\\
            $C[\tau] \gets SR$\\            
    }
    $SR_{new}^i \gets \textbf{mean}(C)$\\
    $S_i \gets  \alpha S_i   + (1-\alpha)  | SR_{new}^i - SR_{old}^i |$\\
    $SR^i_{old} \gets SR_{new}^i$\\
    \If{$SR_{new}^i > max\_SR$}{
        $S_i \gets min\_score$
    }
    \caption{Task sampling estimation using SITP}
\end{algorithm}

\section{Experiments}

In this section, we present an empirical evaluation of SITP approach.
First, we describe the POGEMA environment and show a motivational example with a simple experiment with two tasks. For that experiment, we provide a comparison with the TSCL algorithm and a baseline (uniform sampling of the tasks). Second, we present the results on a number of complex multi-agent pathfinding maps. Finally, we evaluate our approach using Procgen-Benchmark, comparing it with the state-of-the-art curriculum learning technique -- PLR.  

\subsection{Motivational Experiment in POGEMA Environment}

The POGEMA environment~\cite{pogema} is a framework for simulating multi-agent pathfinding problems in partially observable scenarios. Consider $n$ homogeneous agents, navigating the shared map. The task of each agent is to reach the given goal position (grid cell) from the start point in less than $m$ steps. The environment allows both to create maps procedurally and to add existing ones. The environment is multi-agent, thus we consider that the episode ended successfully only in the case when each agent reached its goal. We will refer to this metric as Cooperative Success Rate (CSR).

\begin{wrapfigure}{l}{.5\textwidth}
    \begin{center}
        \vspace{-0.2cm}
        \begin{minipage}[h]{0.48\linewidth}
            \includegraphics[width=1\linewidth]{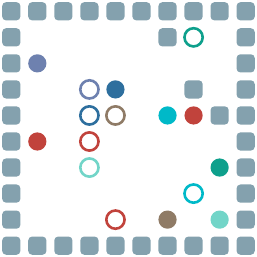} \begin{center} \textbf{(a)} \end{center} 
        \end{minipage}
        \hfill
        \begin{minipage}[h]{0.48\linewidth}
            \includegraphics[width=1\linewidth]{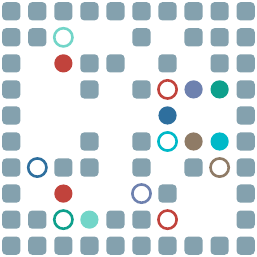} \begin{center} \textbf{(b)} \end{center} 
        \end{minipage}
    \end{center}
    \vspace{-0.2cm}
    \caption{Easy \textbf{(a)} -- easy (obstacle density 5\%) and \textbf{(b)} hard (obstacle density 30\%) Pogema configuration for $8\times8$ map. }
    \label{pic:pogema-env}
\end{wrapfigure}

Even a single task in this domain combines many maps of the same distribution. An example task is a set of maps with size $8\times8$, $16$ agents, and random positions of obstacles with density 30\%, start and goal points.  
We compare SITP with TSCL~\cite{teacher-student}, where the scoring function will take into account the slope of the CSR curve. We use a PPO implementation from Sample Factory paper~\cite{petrenko2020sample}. The hyperparameters were tuned ones on procedurally generated maps without curriculum learning. We use the same network architecture as in~\cite{skrynnik2022pathfinding,ar:pogema}.

\begin{figure}[ht!]
    \begin{center}
        \begin{minipage}[h]{0.49\linewidth}
            \includegraphics[width=1\linewidth]{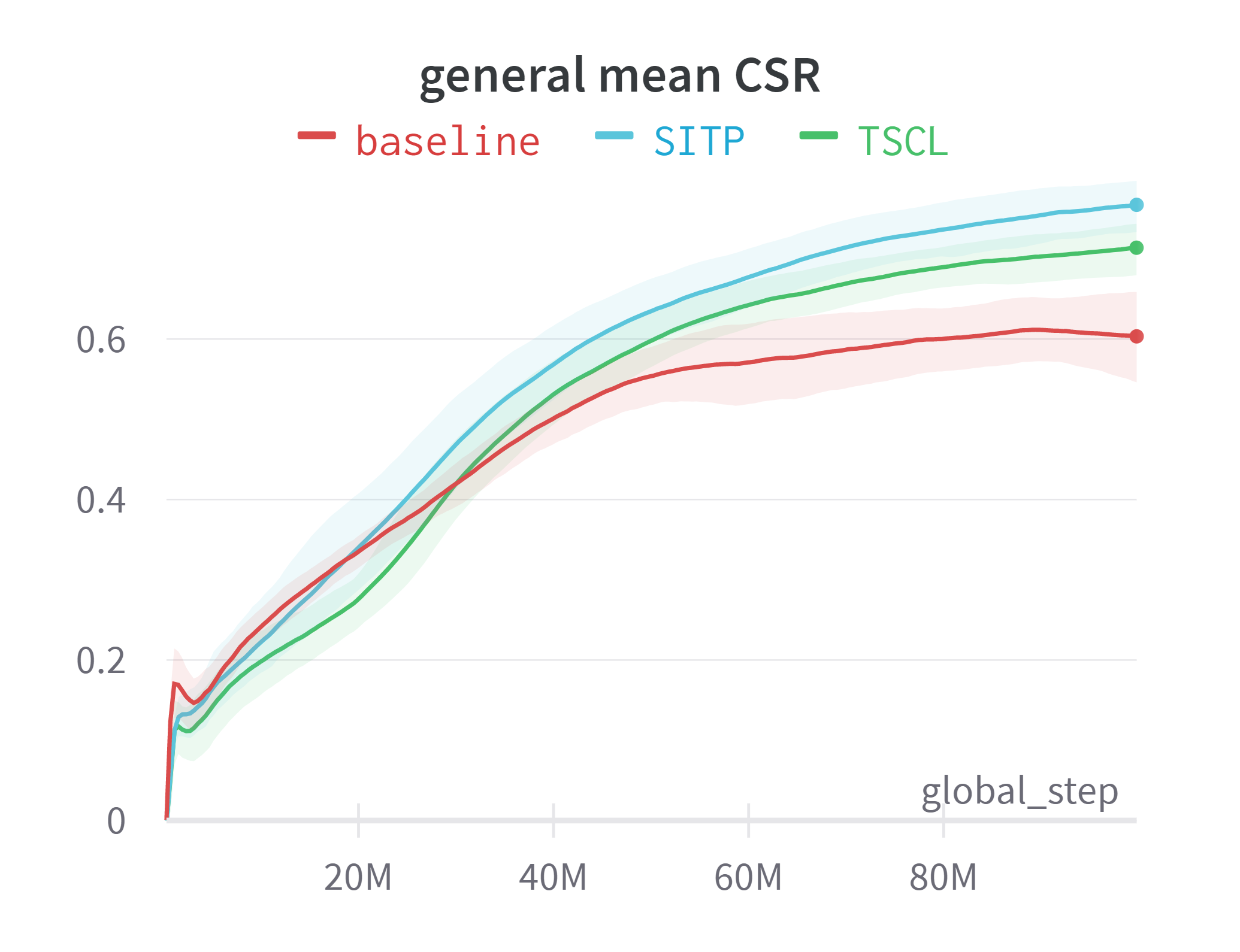} \begin{center} \textbf{(a)} \end{center} 
        \end{minipage}
        \hfill
        \begin{minipage}[h]{0.49\linewidth}
            \includegraphics[width=1\linewidth]{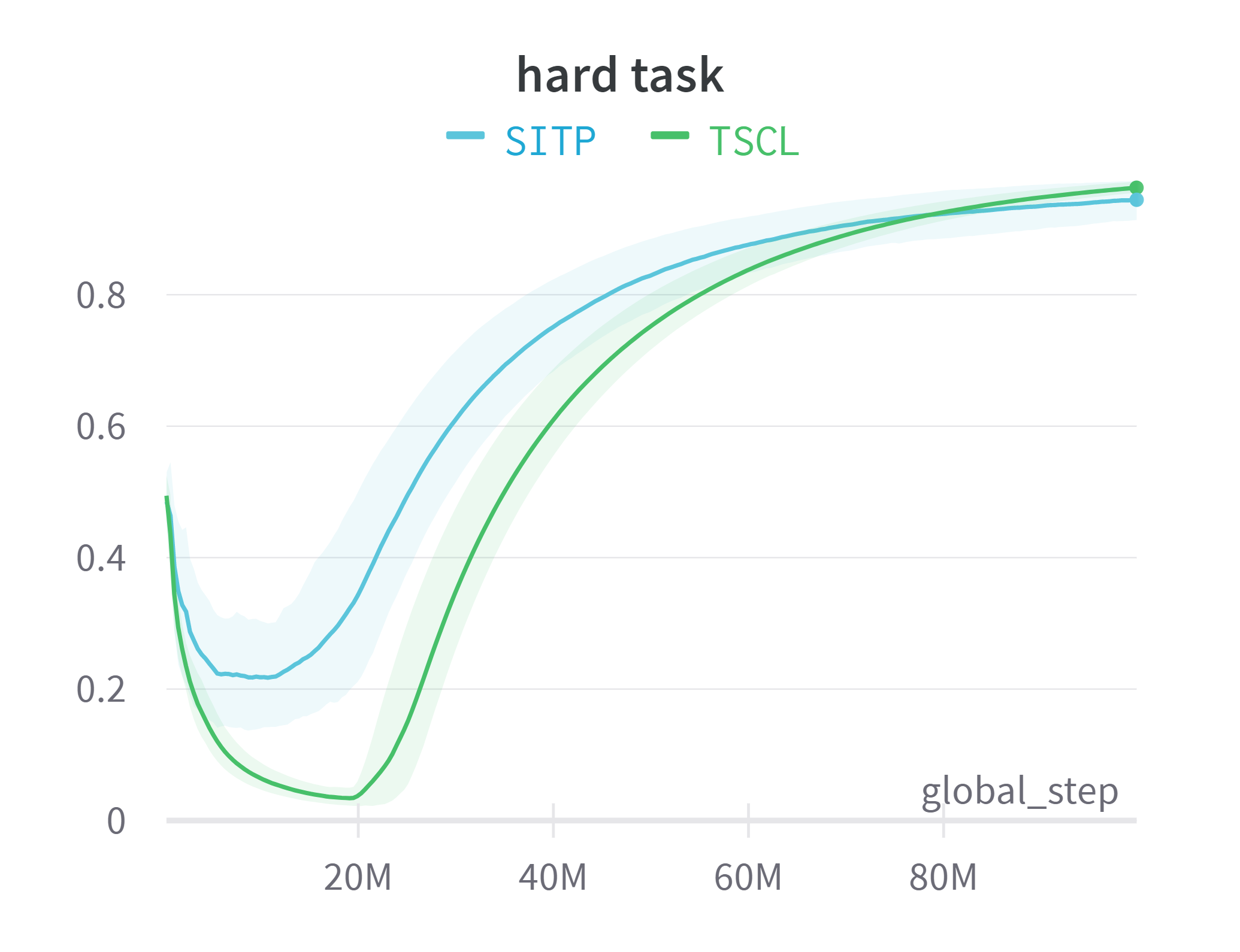} \begin{center} \textbf{(b)} \end{center} 
        \end{minipage}
    \caption{The first experiment is trained on two tasks: easy ($8\times8$ map, $8$ agents, density $5\%$) and hard (8x8 map, 8 agents, density $30\%$). \textbf{(a)} General mean CSR of each method. \textbf{(b)} Percentage of learning on a hard task. The baseline usage of that map is $0.5$, since maps are sampled uniformly. The results are averaged over 10 runs. The shaded area denotes standard deviation.}
    \label{pic:pogema-task2}
    \end{center}
\end{figure}

The first experiment was based on two small tasks, the difference between which was in the complexity of the maps (different density of obstacles). The tasks are shown in Fig.~\ref{pic:pogema-env}. The results are summarized in Fig.~\ref{pic:pogema-task2}. They show that this method quickly learns using the easy task and then selects mostly only the difficult one. SITP outperforms other methods of selecting training levels (TSCL and baseline) in terms of general mean CSR.

\subsection{Large-Scale Experiment in POGEMA Environment}

The second experiment was held on ten large tasks with $64$ agents. Tasks contain maps of different formats: procedurally generated ones (e.g. maps with random obstacles), maps from videogames, maps of warehouses, and indoor maps (e.g. rooms).
We select two procedurally generated configurations: random-64-64-05, random-64-64-3, and $8$ maps from MovingAI dataset~\cite{sturtevant2012benchmarks}: den009d, den204d, den308d, den312d, den998d , room-64-64-8, warehouse-1, warehouse-2.

\begin{figure}[ht!]
    \begin{center}
        \includegraphics[width=0.85\linewidth]{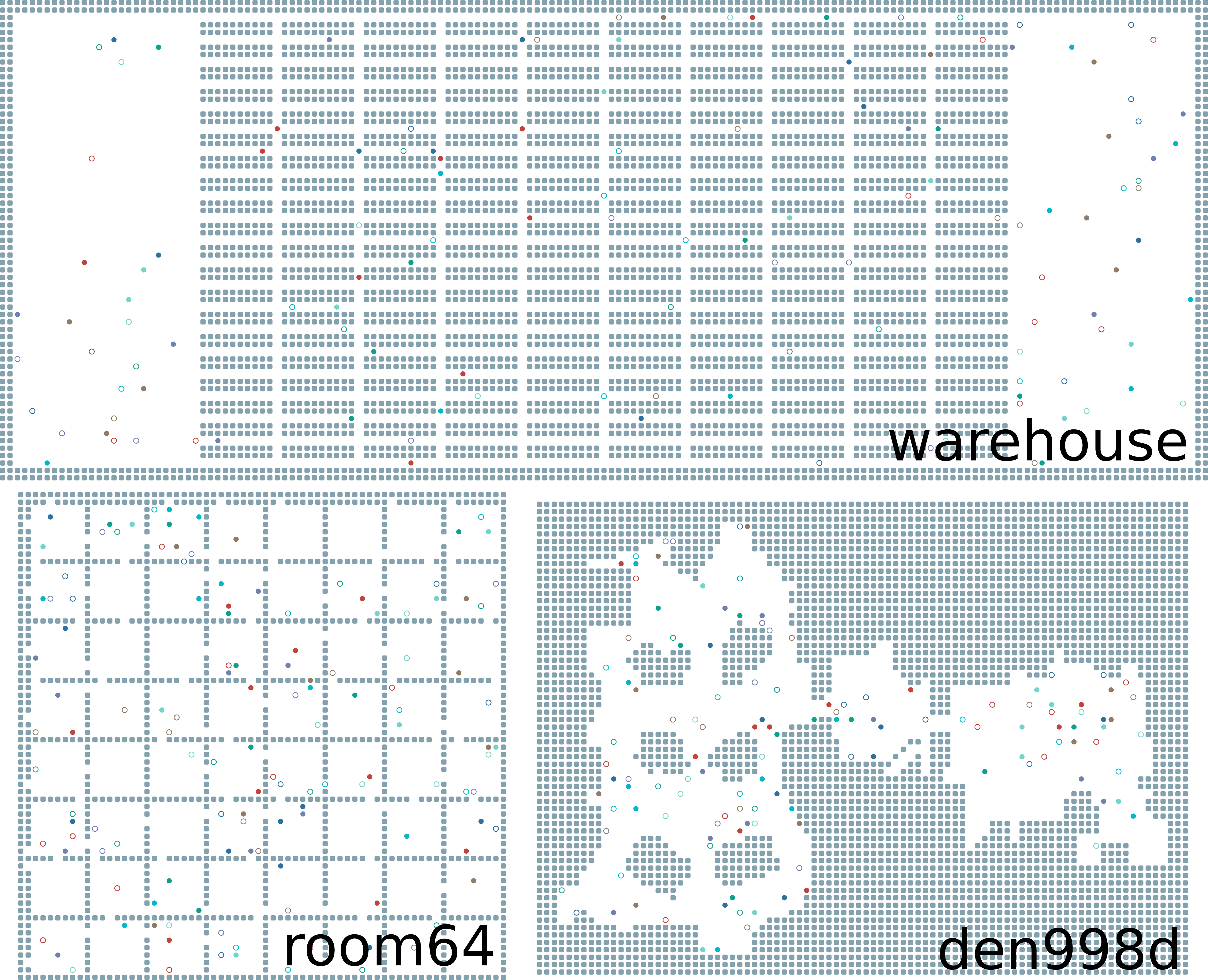}
    \end{center} 
    \caption{Example of three tasks with a fixed obstacle position and with a random position of the start and goal points.}
    \label{pic:pogema-task10}
    \vspace{-0.3cm}
\end{figure}

\begin{figure}[ht!]
    \begin{center}
        \begin{minipage}[h]{0.49\linewidth}
            \includegraphics[width=1\linewidth]{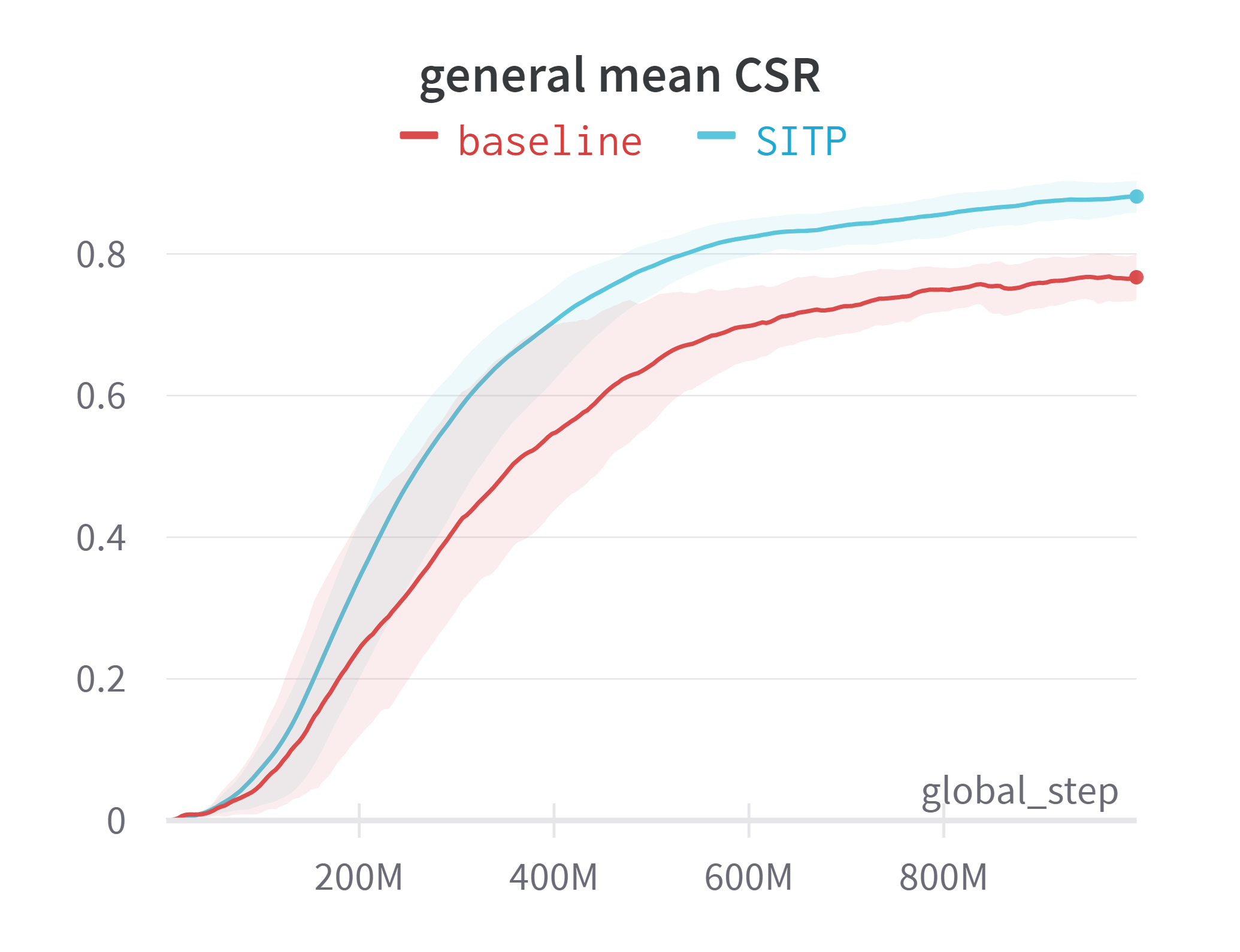} \begin{center} \textbf{(a)} \end{center} 
        \end{minipage}
        \hfill
        \begin{minipage}[h]{0.49\linewidth}
            \includegraphics[width=1\linewidth]{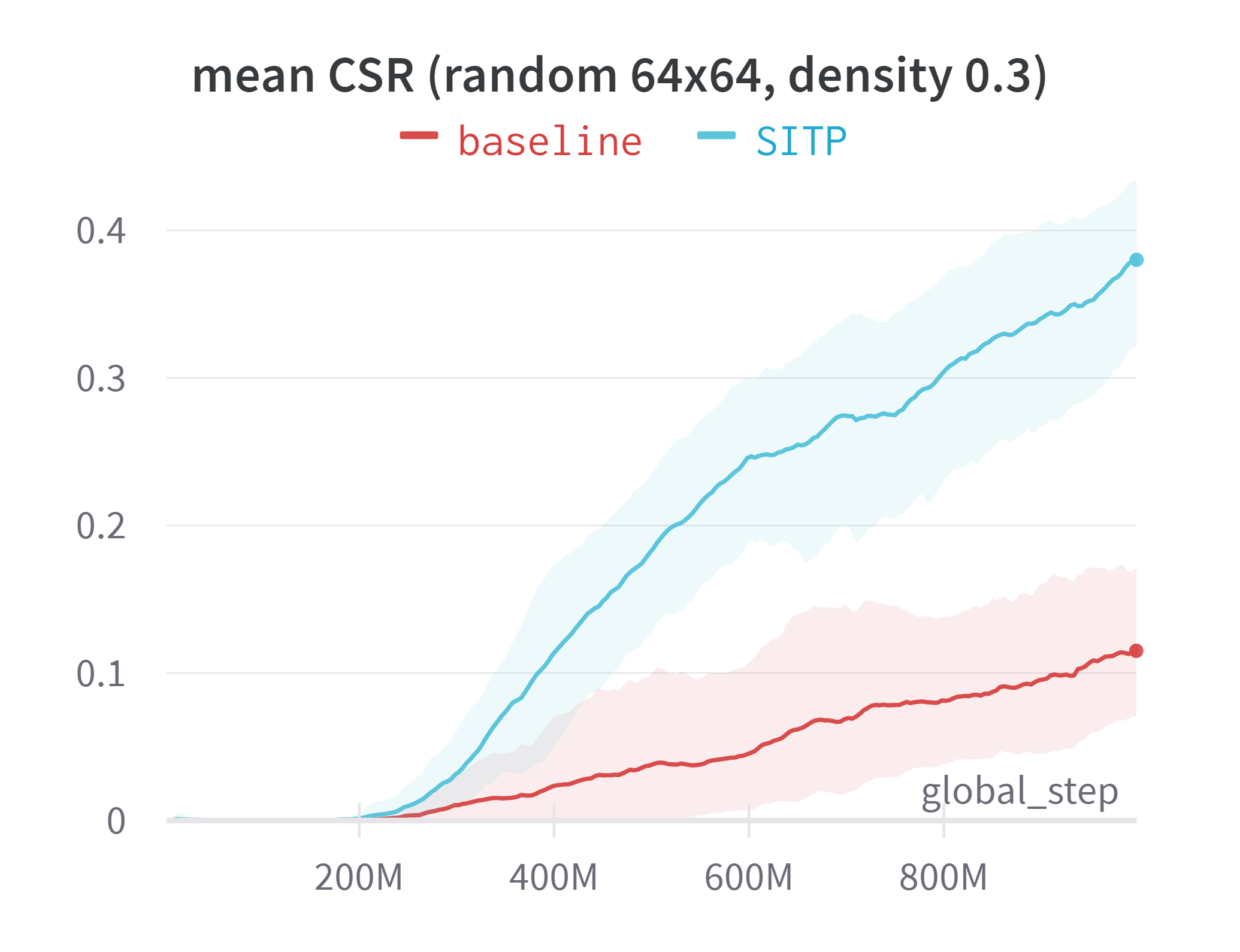} \begin{center} \textbf{\textbf{(b)}} \end{center} 
        \end{minipage}
    \end{center} 
    \caption{Training agents on 10 tasks. \textbf{(a)} General mean CSR of SITP and baseline. \textbf{(b)} Mean CSR for the most difficult procedurally generated task ($64\times64$ map, density 30\%). The results are averaged over 6 runs. The shaded area denotes standard deviation.}
    \label{pic:pogema-2}
\end{figure}

Fig.~\ref{pic:pogema-task10} shows three examples of the maps: \textit{warehouse} is a room where the obstacles are of the same shape and the distance between them is one cell, \textit{room64} are 64 rooms of size 7x7 with at least one exit, \textit{den998d} is a map of a house with obstacles from the videogame. The agent's goal point is generated in such a way so that the agent could always reach it from its starting point (ignoring possible collisions with other agents). Fig.~\ref{pic:pogema-2} shows the superiority of learning results SITP compared to baseline and better results than the TSCL approach. Note that SITP gives a significant increase in mean CSR on hard maps, while mean CSR on easy maps for SITP and baseline are comparable.

\subsection{Procgen Benchmark}

We train and test SITP on $4$ environments in the Procgen Benchmark at simple difficulty levels and make direct comparisons with PLR~ \cite{level-replay}, in which the probability of choosing a task is formed using L1 GAE collected along the last trajectory $\tau$ sampled on that level. The learning potential of a task is calculated using the scoring function based on the GAE magnitude (L1 value loss): $score = \dfrac{1}{T} \sum_{t = 0}^T\left| R_t - V_t\right|$. We reproduce experiments settings from PLR paper and train the agent for $25M$ total steps on $200$ fixed levels, using PPO implementation from that paper. We measure episodic test returns for each game throughout the training. 

For  environments in the Procgen Benchmark, SR is specified by a fixed threshold $SR_{min}$ depending on the total reward per episode such that if $reward > SR_{min}$ then $SR = 1$, otherwise $SR = 0$. $SR_{min}$ is manually selected based on the following rule: the threshold must be greater than the maximum mean episode return per training for baseline, or approximately equal to the maximum mean episode return per training for PLR.


\begin{figure}[ht!]
    \begin{center}
        \includegraphics[width=0.95\linewidth]{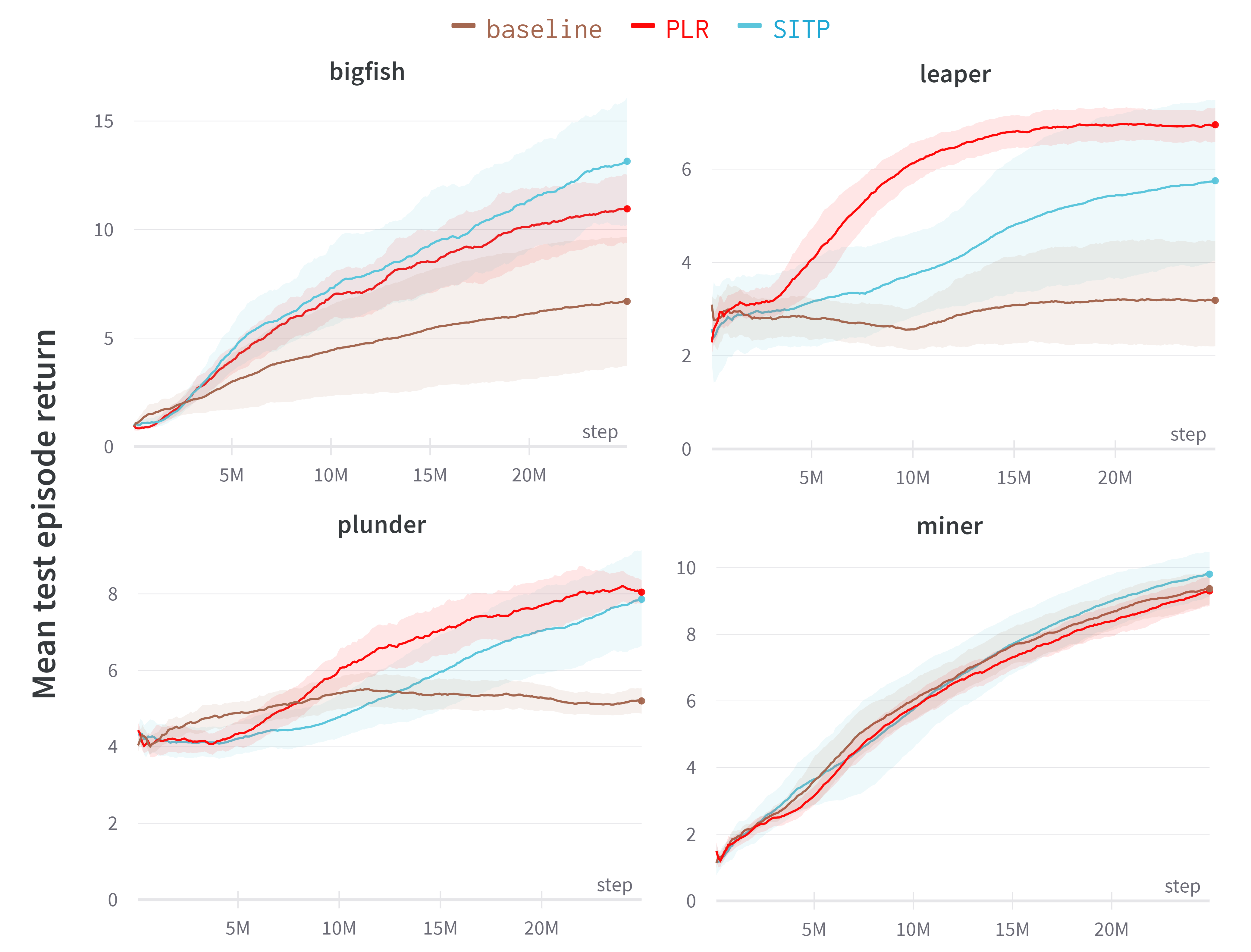}
    \end{center} 
    \caption{Mean episodic test returns over $10$ runs of each method for four environments (bigfish, plunder, leaper and miner). The shaded area denotes standard deviation.}
    \label{pic:procgen}
\end{figure}

In contrast to multi-agent experiments (for which every episode was a new configuration), for Procgen environment we compare agents using evaluation on the test tasks. The results are summarized in Fig.~\ref{pic:procgen}. The fixed threshold was chosen (using expert knowledge) as follows: bigfish ($SR_{min} = 10$), leaper ($SR_{min} = 8$), plunder ($SR_{min} = 12$), miner ($SR_{min} = 10$). We show that SITP, based on the SR, achieves comparable training results with PLR approach. At the same time, PLR higly depends on the implementation of algorithm, and requires additional computations for algorithms differ from Advantage Actor-Critic family.

\section{Conclusion}

In the paper, we investigated the problem of automated curriculum generation for reinforcement learning and introduced Success Induced Task Prioritization (SITP) algorithm, that estimates the learning potential on a task using Success Rate (SR). The SITP can be easily integrated into environments for which a success rate metric is already defined. But also we demonstrated that the SR score can be used in any kind of environments where the reward function is dense and more complicated. 

We showed that SITP improves the efficiency of task sampling in the POGEMA and Procgen Benchmark environments. SITP shows comparable results with the leading curriculum learning methods: TSCL and PLR. Our method does not directly interact with the agent's learning algorithm, it only needs information about the SR after each episode. The future directions of the research includes experiments with other environments. i.e. robotic tasks.

\bibliographystyle{unsrt} 
\bibliography{bib.bib}

\end{document}